\documentclass[preprint]{elsarticle}

\usepackage{hyperref}

\journal{Expert Systems with Applications}

\biboptions{authoryear}

\begin{document}

\begin{frontmatter}

\title{Extraction and Evaluation of Formulaic Expressions Used in Scholarly Papers}

\author[1]{Kenichi Iwatsuki\corref{cor1}}
\ead{iwatsuki@nii.ac.jp}
\cortext[cor1]{Corresponding author}

\author[2]{Florian Boudin}
\ead{florian.boudin@univ-nantes.fr}

\author[3,1]{Akiko Aizawa}
\ead{aizawa@nii.ac.jp}

\address[1]{The University of Tokyo, 7-3-1 Hongo, Bunkyo-ku, Tokyo 113-8656, Japan}
\address[2]{Universit\'e de Nantes, 2 rue de la Houssini\`ere, Nantes 44322, France}
\address[3]{National Institute of Informatics, 2-1-2 Hitotsubashi, Chiyoda-ku, Tokyo 101-8430, Japan}

\begin{abstract}
    Formulaic expressions, such as `\textit{in this paper we propose}', are helpful for authors of scholarly papers because they convey communicative functions; in the above, it is `\textit{showing the aim of this paper}'.
    Thus, resources of formulaic expressions, such as a dictionary, that could be looked up easily would be useful.
    However, forms of formulaic expressions can often vary to a great extent.
    For example, `\textit{in this paper we propose}', `\textit{in this study we propose}' and `\textit{in this paper we propose a new method to}' are all regarded as formulaic expressions.
    Such a diversity of spans and forms causes problems in both extraction and evaluation of formulaic expressions.
    In this paper, we propose a new approach that is robust to variation of spans and forms of formulaic expressions.
    Our approach regards a sentence as consisting of a formulaic part and non-formulaic part.
    Then, instead of trying to extract formulaic expressions from a whole corpus, by extracting them from each sentence, different forms can be dealt with at once.
    Based on this formulation, to avoid the diversity problem, we propose evaluating extraction methods by how much they convey specific communicative functions rather than by comparing extracted expressions to an existing lexicon.
    We also propose a new extraction method that utilises named entities and dependency structures to remove the non-formulaic part from a sentence.
    Experimental results show that the proposed extraction method achieved the best performance compared to other existing methods.
\end{abstract}


\end{frontmatter}

\section{Introduction}

Writing scientific papers is crucial but a laborious task in research activities, especially for non-native English speakers.
\citet{Zhao2017} and \citet{Wu2020} demonstrated that the quality of English academic writing is significantly different between native and non-native researchers.
Also, it is time-consuming to look up words in a dictionary or ask for English proofreading.
Thus, writing assistance can be a great help to non-native researchers to improve the quality of their papers and to save much time in writing, which will accelerate their research activities.

As a means of writing assistance, the use of \textit{formulaic expressions} has previously been investigated \citep{AlHassan2015,Mizumoto2017,Iwatsuki2018}.
Formulaic expressions are continuous or discontinuous word sequences that are frequently used in scientific papers to convey specific \textit{communicative functions} \citep{Cortes2013,Adel2014}.
For example, the formulaic expression `\textit{little attention has been paid to}' conveys the communicative function `\textit{referring to the paucity of past work}'.
Instead of having to compose everything by themselves, the use of formulaic expressions helps authors express their intended meaning more properly and effectively. 

To utilise them, formulaic expressions should first be collected from a corpus of scientific papers.
However, the difficulty lies in both automatic extraction of formulaic expressions and automatic evaluation of formulaic expressions.
In previous studies \citep{Hyland2008,Chen2010,Simpson-vlach2010}, frequent word $n$-grams have been extracted from a corpus and the usefulness of extracted word sequences has been evaluated manually because of a lack of automatic evaluation methods.
However, formulaic expressions are not always fixed lexical units.
Some words can be replaced with others and spans are also flexible.
For example, `\textit{in this paper we propose}' is a formulaic expression, but `\textit{in this study we propose}' and `\textit{in this work we propose}' sometimes appear instead.
Also, both `\textit{in this paper we propose}' and `\textit{in this paper we propose a new method to}' can be regarded as formulaic expressions because they both convey the communicative function `\textit{showing the aim of the paper}'.
However, `\textit{paper we propose a}' should not be labelled as a formulaic expression.
In short, forms of formulaic expressions can vary according to the syntax and content of the sentence in which they appear.
Therefore, the existing approach has made it difficult to automatically determine which word sequences should be formulaic expressions.

\begin{figure}[t]
    \centering
    \includegraphics[scale=0.7]{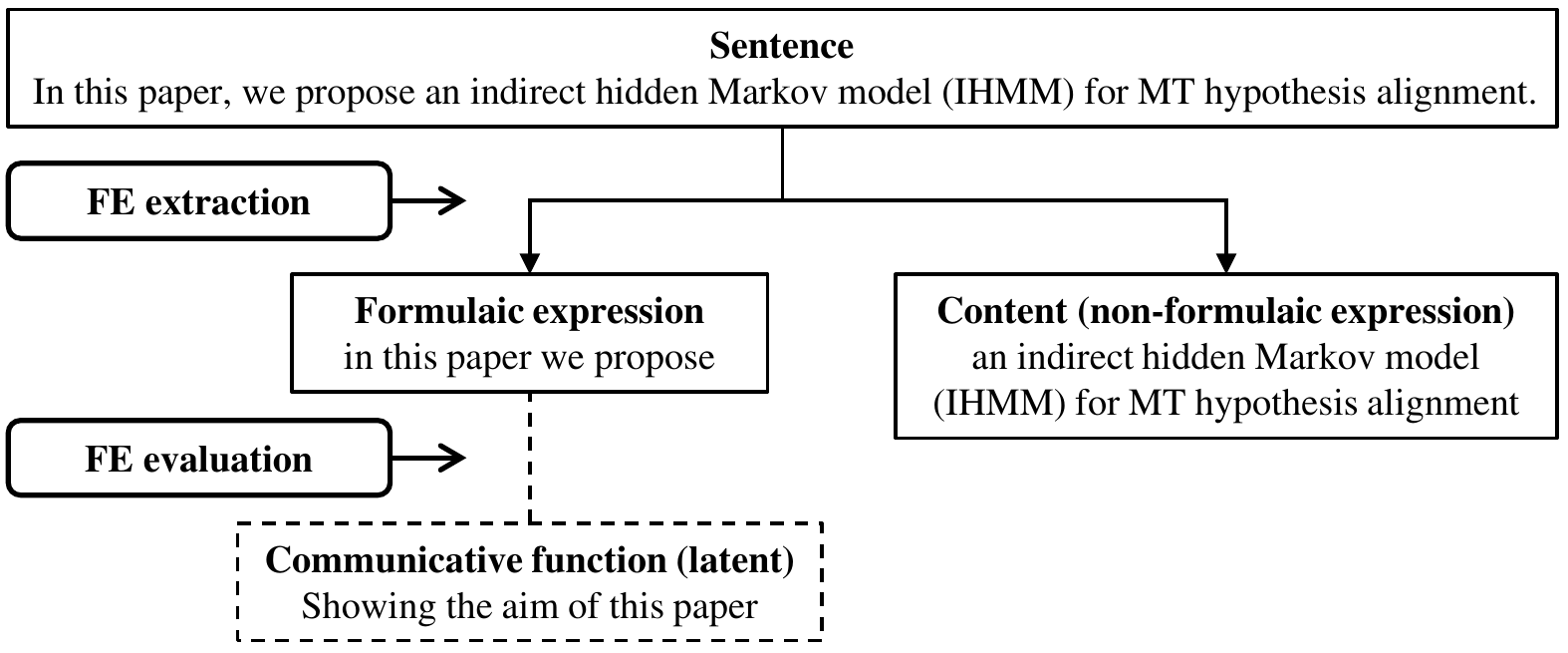}
    \caption{Sentence from a paper \citep{He2008} presented in ACL Anthology. We assume that a sentence consists of a formulaic expression that conveys a specific communicative function and content. Thus, extraction of formulaic expressions is to distinguish formulaic part from the non-formulaic part of a sentence. Also, to evaluate the extraction methods, how strongly a formulaic expression and communicative function are connected is measured.}
    \label{fig:disentanglement}
\end{figure}

To solve these problems, we redefine the extraction and evaluation problems in the following way.
First, formulaic expressions are always used in a sentence, never alone.
Therefore, we assume that a sentence consists of two parts: a formulaic expression that conveys a specific communicative function and a remaining non-formulaic part that expresses content such as names of materials and details of methods (Figure~\ref{fig:disentanglement}).
From this viewpoint, the extraction task can be regarded as a sequential labelling problem, that is, labelling each word in a sentence formulaic or non-formulaic.
For evaluation we measure how strongly connected are an extracted formulaic expression and a communicative function.
Unlike previous methodologies, which focus only on formulaic expressions rather than whole sentences, our approach makes it possible to deal with short, long, frequent and infrequent formulaic expressions at once.

\begin{figure}[t]
    \centering
    \includegraphics[scale=0.6]{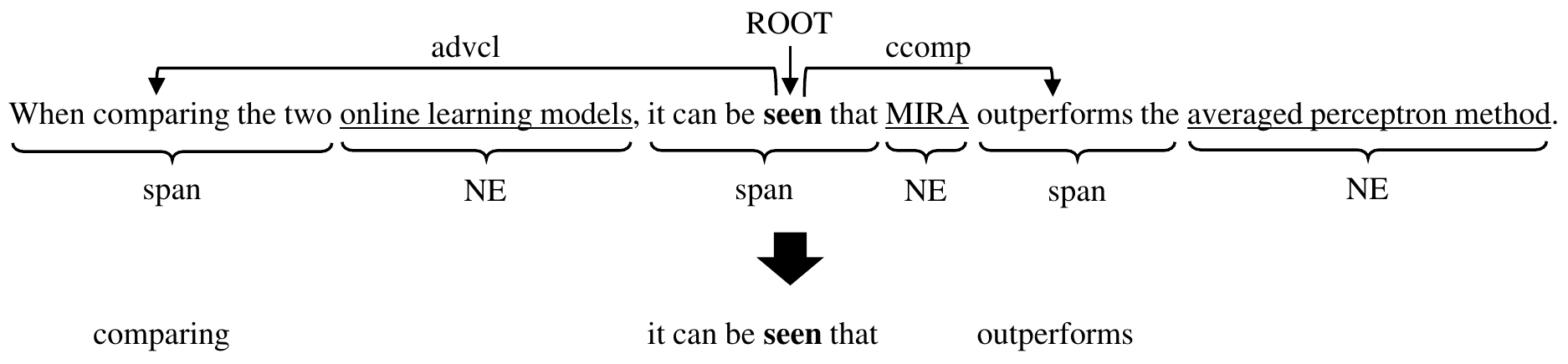}
    \caption{We first remove named entities (NE) from a sentence, resulting in three spans in this example. Then, we remove words not satisfying the two conditions: (1) all the words in the span that contains a root \textit{and} (2) words organised by a root.}
    \label{fig:propsoed_method_image}
\end{figure}

Additionally, based on this approach, we propose an extraction method that utilises named entities and a dependency structure to remove the non-formulaic part from a sentence (Figure~\ref{fig:propsoed_method_image}).
First, we remove named entities in a sentence, resulting in a few spans split by the named entities.
Secondly, we select words to remove based on the dependency structure of the sentence.
Words that do not belong to a span containing the root of the sentence and that are not organised by the root are removed.

For evaluation, we also measure how much a formulaic expression conveys a communicative function by assigning different weights to formulaic and non-formulaic words in a sentence.
To do so, we propose using the sentence retrieval task \citep{Iwatsuki2020} as an extrinsic evaluation method.
In this task, a query sentence is given and sentences that have the same communicative function as the query should be retrieved.
Sentences are converted into vector representations and ranked according to their similarity with the query.
Each sentence is tagged with its communicative function in advance.
The difference between the original task and our evaluation task lies in how the sentence vectors are created.
In the original setting, sentence vectors are created by averaging vectors of each word in a sentence, which is a well-known way to create them.
On the contrary, to examine how much the formulaic part of a sentence conveys a communicative function, we propose creating sentence vectors by assigning different weights to formulaic words and non-formulaic words in a sentence.

We compare the performance of our proposed method to that of existing extraction methods.
The results show that the proposed method achieves the best performance among all compared methods.

Our contributions are as follows.
First, we propose a comprehensive approach to extract and evaluate formulaic expressions that can take a variety of forms.
Secondly, we propose a new method to evaluate extraction methods by assigning different weights to formulaic expression candidates to create sentence representations and applying a sentence retrieval task as an extrinsic evaluation.
Thirdly, we empirically demonstrate that the proposed evaluation method is valid by testing formulaic and non-formulaic expressions.
Finally, we propose a new method to extract formulaic expressions. We empirically verified that the proposed method achieves the best performance among all the methods we tested.

The proposed method does not require additional data labelled with formulaic expressions and it can be immediately applied to other corpora.
Thus, this work will accelerate the construction of multi-disciplinary database of formulaic expressions and research on computer-aided writing assistance using formulaic expressions.
Moreover, because formulaic expressions are used not only in scholarly papers but also in other documents and speeches, we hope the present study can contribute to enhancing writing communications.

\section{Related Work}

\subsection{Communicative Functions in Scholarly Papers}

Communicative functions represent the intentions of authors of scholarly articles.
Authors must communicate with readers in order for them to understand their research properly.
Thus, every part of a scientific paper has a specific function, such as providing background information, explaining methodology and discussing experimental results, and readers interpret these functions to understand why that text is written.

\begin{figure}[t]
    \centering
    \includegraphics[scale=0.75]{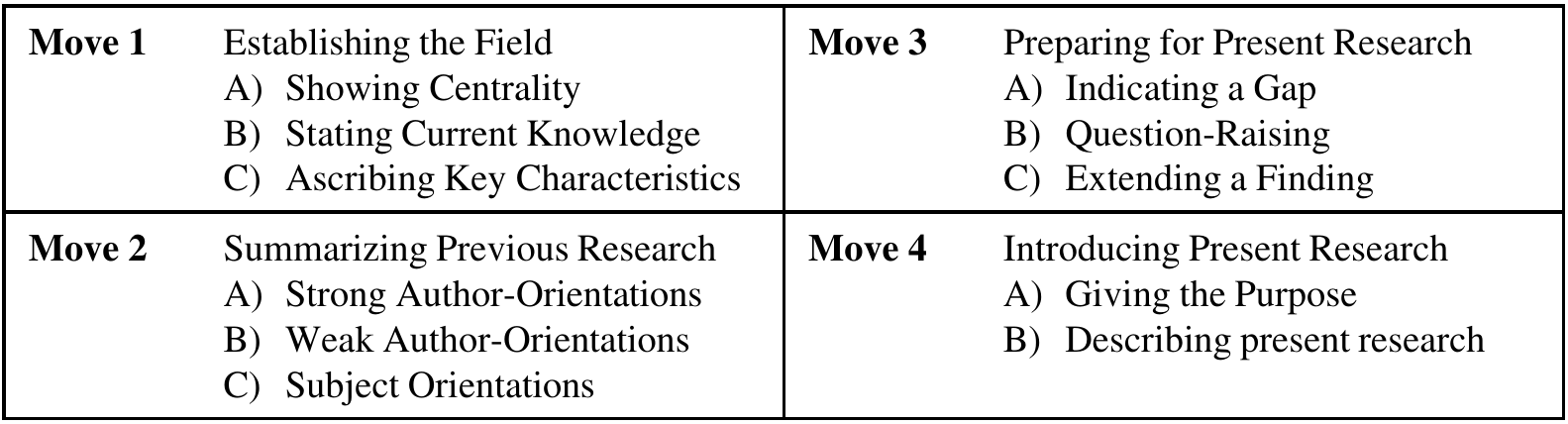}
    \caption{Moves in introduction sections proposed by \citet{Swales1981}. There are four moves appearing in this order in the section. Each move has two or three steps, which are finer-grained communicative functions.}
    \label{fig:swales4moves}
\end{figure}

Communicative functions should be aligned in a reasonable order that is conventionally established by the research community to make papers easily understandable.
\citet{Swales1981} first introduced the concept of \textit{move}, which is a rhetorical unit conveying a communicative function in scholarly papers.
Transitions of moves have been found to be fixed to some extent.
In Figure~\ref{fig:swales4moves} moves and their transitions in introduction sections are described.
Each move has several \textit{steps}, denoted by A), B) and C), which are finer-grained units.
Following his work \citep{Swales1981,Swales1990,Swales2004}, which focused on the introduction sections in research articles, \citet{Cotos2015} and \citet{Maswana2015} analysed moves in every section.
They created lists of moves and steps found in scholarly articles.

Units where communicative functions are realised are flexible.
Several sentences sometimes realise one communicative functions, while a clause may also do.
However, in previous work \citep{Hirohata2008,Dayrell2012,Fiacco2019,Iwatsuki2020}, a sentence was regarded as a unit of communicative function.
We follow this manner; we assume that one sentence has a communicative function and thus one sentence has one formulaic expression that conveys the communicative function.

There are a few studies dealing with classification of communicative functions.
\citet{Dayrell2012} and \citet{Hashimoto2016} proposed feature-based machine learning methods to classify sentences according to their communicative functions.
The limitation of these studies is that they used only abstracts of papers.
Thus, classification of communicative functions of a whole paper remains an open issue.

\subsection{Formulaic Expressions and Communicative Functions for Writing Assistance}

Formulaic expressions used in academic writing, also known as formulaic sequences, lexical bundles and phraseologies, have been studied by many researchers \citep{Simpson-vlach2010,Adel2012,Liu2012,Vincent2013,Perez-Llantada2014,Omidian2018}.
The usage of formulaic expressions differs across disciplines \citep{Hyland2008,Nekrasovabeker2019}.
Domain-specific studies on formulaic expressions, including mathematics \citep{Cunningham2017}, social sciences \citep{Lu2018}, medicine \citep{Jalali2014}, psychology \citep{Esfandiari2017} and applied linguistics \citep{Qin2014}, have been conducted.
Therefore, not only general-purpose formulaic expressions but also domain-specific formulaic expressions should be collected for writing assistance.

\citet{Cortes2013} and \citet{Adel2014} proposed combining formulaic expressions and communicative functions.
This combination makes it relatively easy to search for specific formulaic expressions because formulaic expressions labelled with their communicative functions can be searched for by not only keywords but also authors' intentions.
Thus, a recently proposed writing assistance system adopts this approach \citep{Mizumoto2017}.
Following these studies, in this work, we adopt the definition that formulaic expressions are combined with communicative functions.

\subsection{Multi-Word Expressions and Formulaic Expressions}

Generally, multi-word expression is a different concept to formulaic expression but there is some overlap between the two concepts.
Multi-word expressions do not always convey a communicative function.
According to the survey by \citet{Constant2017}, multi-word expressions can be categorised in several ways.
For instance, `\textit{kick the bucket}' is a typical multi-word expression and categorised into the \textit{idiom} class and `\textit{International Business Machines}' is categorised into the \textit{multi-word named entity} class.
However, both do not convey any specific communicative function in scientific papers.

PARSEME \citep{Savary2017} is the most comprehensive dataset for multi-word expression identification.
In this dataset, multi-word expressions are classified into three categories: general, quasi-general and other; these categories are not based on communicative functions.
Therefore, state-of-the-art models for identification of multi-word expressions trained on the dataset \citep{Waszczuk2019,Saied2019} cannot be directly applied to the extraction of formulaic expressions.

\subsection{Evaluation of Formulaic Expressions}

Manual evaluation has been a common method of formulaic expression evaluation.
\citet{Simpson-vlach2010} asked experts whether they thought extracted formulaic expressions were formulaic or had cohesive meaning and \citet{Iwatsuki2018} asked annotators whether they thought extracted formulaic expressions were helpful for writing.
Generally speaking, for tasks of building new vocabulary, there is no reference.
If some reference data exist, we do not need to create another, which \citet{Brooke2015} also pointed out.
Thus, an automated evaluation in which all extracted candidates are compared to a reference lexicon is not realistic.

Additionally, the flexibility of formulaic expressions also makes automated intrinsic evaluations difficult, where extracted formulaic expression candidates are evaluated by their properties, such as frequency and mutual information.
For example, both `\textit{beyond the scope}' and `\textit{is beyond the scope of this paper}' are good formulaic expressions that convey the same communicative function, i.e., `\textit{describing the limitations of current research}'.
Therefore, even if manually annotated formulaic expressions are available, there are still other allowable formulaic expressions as long as they convey the same communicative function.

To avoid these problems, we first propose an extrinsic evaluation method that utilises communicative functions conveyed by formulaic expressions.
The idea is that a sentence can be split into a formulaic expression and a content part and the former should convey a communicative function.
Therefore, how strongly a formulaic expression candidate is connected to a sentence's communicative function can be considered a good proxy for measuring of the quality of the formulaic expression candidate.
We adopt the communicative-function-oriented sentence retrieval task to check the degree of the connections.

\section{Methods}

\subsection{Dataset}

We use two datasets for different purposes.
The first dataset is the ACL Anthology Sentence Corpus (AASC)\footnote{\url{https://github.com/KMCS-NII/AASC}}, which consists of 13,923 papers retrieved from ACL Anthology\footnote{\url{https://www.aclweb.org/anthology/}}.
For each paper, narrative texts are split into sentences and sentences are labelled with their section.
Generally, section headers in papers are not always fixed to a set of labels such as introduction, methods, results and discussion, even though the content of the sections can be classified into these fixed categories.
For example, there is a case where two sections of two different papers explain methodologies but the section headers are different: `Learning Method' and `Approach'.
Thus, it is necessary to integrate these variants into one content-based section header, i.e.,  `methods' in this example.
However, in this dataset, the section labels are normalised into a limited number of labels; thus, we can use sentences without checking the original section titles.

\begin{figure}[t]
    \centering
    \includegraphics[scale=0.7]{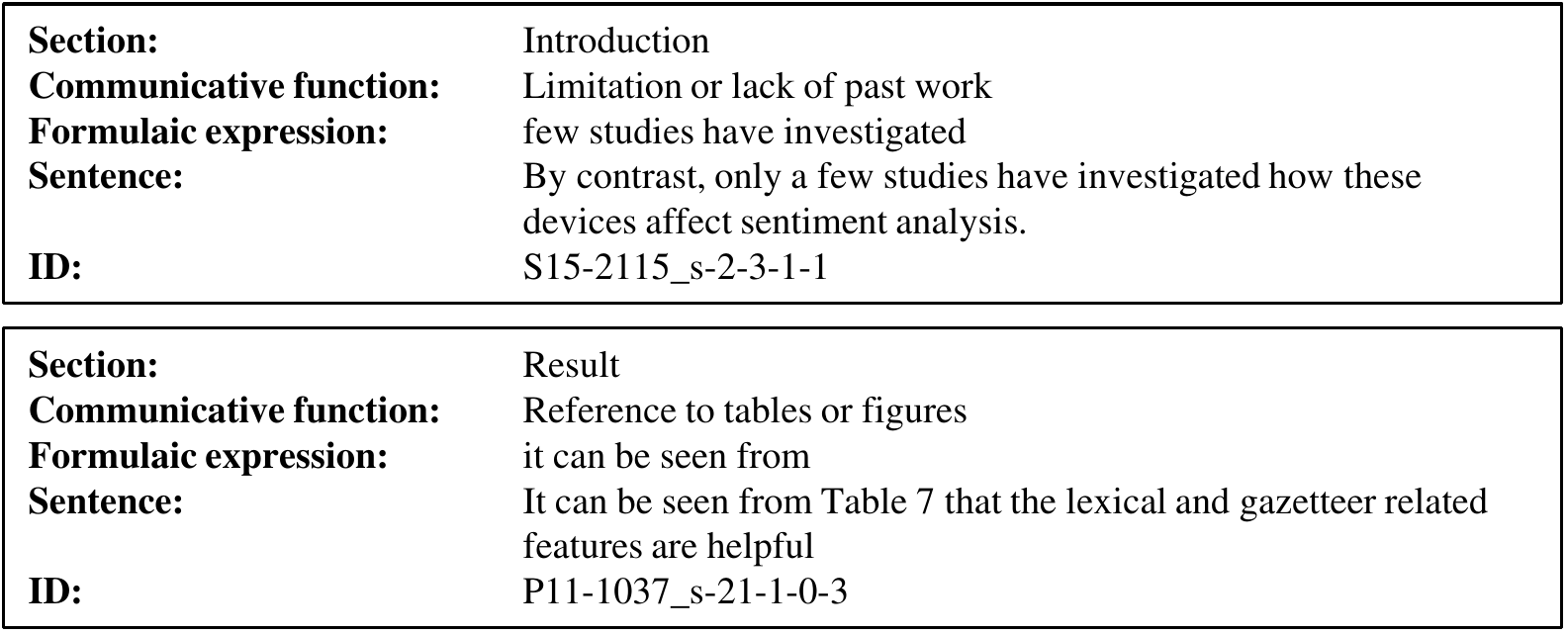}
    \caption{Two examples recorded in the FECFeval dataset. Each instance consists of a section label, communicative function, formulaic expression and sentence. These sentences were originally retrieved from \citet{Hee2015} and \citet{Liu2011}.}
    \label{fig:FECF}
\end{figure}

The second dataset (FECFeval)\footnote{\url{https://github.com/Alab-NII/FECFevalDataset}} created by \citet{Iwatsuki2020} consists of 5 sections (introduction, background, method, result and discussion).
Each instance in the dataset consists of a sentence extracted from AASC, annotated with its communicative function and formulaic expression (see examples in Figure~\ref{fig:FECF}).
The number of communicative functions is 39: 11 for introduction, 7 for background, 6 for method and result and 9 for discussion; the total number of instances is 691.
The communicative functions are based on the existing resource, Academic Phrasebank\footnote{John Morley, \url{http://www.phrasebank.manchester.ac.uk/}}.

\subsection{Extraction}

We assume that a sentence consists of a formulaic expression that conveys a communicative function and named entities that realise a content of a sentence\footnote{Of course, there are sentences that do not contain formulaic expressions but this task is the extraction of formulaic expressions; thus, we focus only on sentences containing formulaic expressions. Also, some sentences do not contain any named entities but this method can still be applied; nothing will be removed from a sentence.}.
Therefore, instead of directly identifying the formulaic part, we apply named entity recognition (NER) to remove the content part from a sentence.
We also investigated how many manually annotated formulaic expressions in the FECFeval dataset contain words that are roots in the sentence dependency structure and we found that 442 out of 686 (64.4\%) formulaic expressions contain roots.
Thus, we extract a root of a sentence using the dependency structure of a sentence.

Named entity removal is conducted in the following way.
In a sentence, there can be both named entities specific to scientific papers, such as names of methods, and datasets and general named entities, such as locations.
Thus, we use two different datasets to train the NER model: SciERC \citep{Luan2018} and CoNLL04 \citep{Roth2004}.
SciERC is a dataset based on scholarly papers and named entities are annotated.
Its entity types are specific to scientific papers: task, method, evaluation metric, material, other scientific terms and generic.
CoNLL04's entity annotations are general ones: location, organisation, people and other.
The NER model we trained on the two datasets is SpERT\footnote{We used the implementation presented by the authors: \url{https://github.com/markus-eberts/spert} .} \citep{Eberts2020}, which is the top of the leader board of NER tasks in SciERC\footnote{Spert achieves the best performance on NER on SciERC according to `paper with code' (\url{https://paperswithcode.com/sota/named-entity-recognition-ner-on-scierc}) as of 12 April 2020.}.

By the removal of named entities, a sentence can be split into several spans (if no named entity is in a sentence, no split happens).
We applied the Stanford CoreNLP dependency parser \citep{Qi2018} to remove words that did not belong to a span containing a root and were not organised by a root.

\begin{figure}[t]
    \centering
    \includegraphics[scale=0.5]{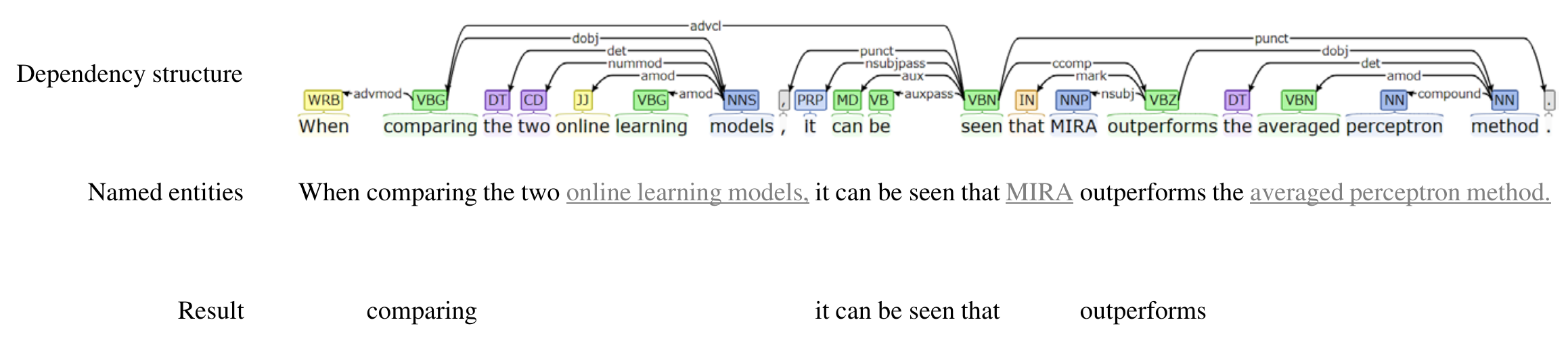}
    \caption{Result of dependency parsing and named entity recognition. Named entities are coloured grey and underlined.}
    \label{fig:NERdepparse}
\end{figure}

In Figure~\ref{fig:NERdepparse}, an example of a sentence processed by NER and dependency parsing is shown.
In this example, named entity removal results in three spans: `when comparing the two', `it can be seen that' and `outperforms the'.
The root of this sentence is `seen'; thus, the span `it can be seen that' was marked as the formulaic part.
Additionally, the words in the other spans that are organised by the root, namely `comparing' and `outperforms', remained.
All the other words were dropped; then, the formulaic expression candidate is `comparing * it can be seen that * outperforms'.

\subsection{Evaluation}

\subsubsection{Sentence Representations}

As mentioned in the introduction, we assume that a communicative function is conveyed by a formulaic expression and thus, the extraction can be evaluated by the strength of connection between a formulaic expression and a communicative function.
Therefore, we create sentence vectors by assigning different weights to the formulaic and non-formulaic parts.
It is a common way to average word embeddings of each word of a sentence to create a sentence vector.
Unlike the ordinary method, we assign different weights to word vectors of formulaic and non-formulaic parts when averaging them, which can be formalised as follows:
\begin{eqnarray}
    \mathrm{s}(W) = \frac{1}{|W|} \Biggl\{ \alpha \cdot \sum_{w_i \in \mathrm{FE}} \mathrm{v}(w_i) + (1-\alpha) \cdot  \sum_{w_j \in \mathrm{nonFE}} \mathrm{v}(w_j) \Biggr\}, \nonumber 
\end{eqnarray}
where $\mathrm{s}(\cdot)$ is a vector of a sentence, $W$ is a sequence of words in the sentence, which consists of $\mathrm{FE}$ (formulaic expression) and $\mathrm{nonFE}$ (the remaining words in the sentence), $\mathrm{v}(w)$ is a function that returns a vector representation of $w$ and $\alpha (0 \leq \alpha \leq 1)$ is a parameter determining the weights of the formulaic and non-formulaic parts.
When $\alpha=0.5$, the sentence vector is simply the average of each word embedding.
When $\alpha=1.0$, it consists of only the formulaic part.

Unlike the experiments conducted in \citet{Iwatsuki2020}, where $\alpha$ was fixed to 0.5, we vary $\alpha$.
In our experimental setting, we use skip-gram models for $\mathrm{v}(w)$ trained on AASC. We follow the experimental settings used in \citet{Iwatsuki2020}: the dimension is 200 and the window size is 2.
It should be noted that our experiments do not rely on specific word embedding models or parameters.

\subsubsection{Sentence Retrieval Task}

Instead of directly evaluating extracted formulaic expressions, we propose an extrinsic evaluation method that utilises communicative functions conveyed by formulaic expressions.
We adopt the sentence retrieval task proposed by \citet{Iwatsuki2020} to measure the strength of connection between extracted formulaic expressions and communicative functions.
In this task, a query sentence is given and then a retrieval system should return an ordered list of sentences ranked according to the similarities of communicative functions between the query and other sentences.
Then, the top-$N$ sentences in the list are selected and for evaluation, it is checked how many sentences have the same communicative function as the query.

In the system, sentences are converted into vector representation, as described above.
Then, sentence vectors are ranked according to the cosine similarity with the query.
Mean average precision (MAP) is used for evaluation of the retrieval task, which is formulated as follows:

\[
        \mathrm{MAP}(S^i) = \frac{1}{|S^i|} \sum_{s_j \in S^i} \frac{1}{n_{s_j}} \sum_{k=1}^{|R_j^i|}
    \left\{
        \begin{array}{ll}
            0 & (\mathrm{CF}(r_k) \neq \mathrm{CF}(s_j)) \\
            \mathrm{P}_j^i(k) & (\mathrm{CF}(r_k) = \mathrm{CF}(s_j))
        \end{array},
    \right.
\]
where $S^i$ is a set of sentences in section $i$, $n_{s_j}$ is the number of correct answers when the query sentence is $s_j$, $R^i_j$ is an ordered list of the sentence retrieval result, $\mathrm{P}_j^i(k)$ is the precision at position $k$-th in the list and $\mathrm{CF}(r_k)$ is a communicative function of the $k$-th ranked sentence $r_k \in R_j^i$.

\section{Experiments}

\subsection{Overview}

We conducted two experiments.
The first one is for validating whether our proposed evaluation method works or not.
We prepared manually annotated formulaic and non-formulaic expressions and compared their performances in sentence retrieval.
The second one compared our proposed extraction method to other existing methods.

Both experiments are proceeded in the following way.
First, the FECFdataset was split into five sections (introduction, background, method, result and discussion).
Secondly, for each section, one sentence was chosen as a query, and the sentence retrieval was applied to a set of other sentences.
Then, another sentence in the section was chosen as a query, and the same process was repeated.
After all the sentences were used as a query, the MAP score for the section was calculated.
Finally, the average of all five MAP scores was calculated for evaluation.
For simplicity, we refer to the averaged MAP score as MAP score hereafter.

\subsection{Validity of the Evaluation Method}

In the FECFeval dataset \citep{Iwatsuki2020}, the \textit{CoreFEs} are labelled for each sentence.
CoreFEs are phrases that are manually labelled as formulaic expressions that convey a specific communicative function, but only the core part of a formulaic expression is annotated because CoreFEs are used as query keywords for the retrieval of sentences from a corpus, in which a query that is too long would result in no matching results. 
For example, `\textit{to the best of our knowledge no work exists on}' can be regarded as a formulaic expression but `\textit{no work exists}' is only labelled as a CoreFE.
Thus, it should be noted that a CoreFE can be regarded as a formulaic expression but it misses some words that could also be included in the formulaic expression.
We used the CoreFEs as the result of manual extraction to compare other methods of extraction. 

\begin{figure}[t]
    \centering
    \includegraphics[scale=0.5]{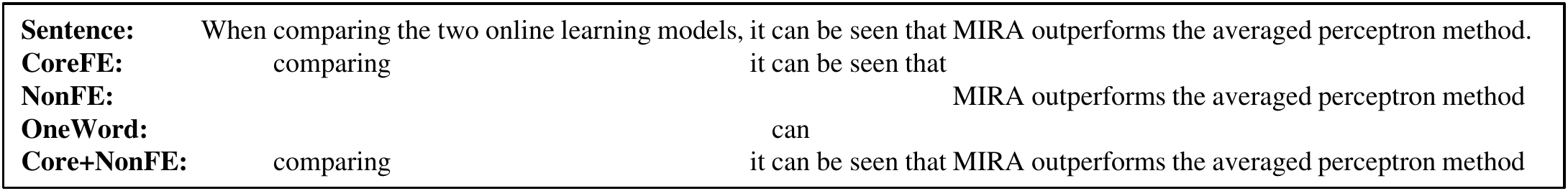}
    \caption{Examples of four methods: CoreFE, NonFE, OneWordCoreFE (OneWord) and CoreFE+NonFE (Core+NonFE), all of which are extracted from the sentence.}
    \label{fig:CoreFE_random_example}
\end{figure}

For comparison purposes, we prepare three other types of expressions: NonFE, OneWordCoreFE and NonFE+CoreFE.
Figure~\ref{fig:CoreFE_random_example} shows the examples of the four patterns.
NonFE represents words that are randomly extracted from a sentence in which a CoreFE is removed.
The length of NonFE expressions is the same as that of the corresponding CoreFE.
These are regarded as bad formulaic expressions.
OneWordCoreFE represents one word randomly picked from a CoreFE for each sentence.
NonFE+CoreFE represents combinations of NonFE and CoreFE.

OneWordCoreFE simulates an extraction method that misses most parts of formulaic expressions.
Putting more weight on OneWordCoreFE means applying less weight to most parts of formulaic expressions.
Thus, the performance should start to deteriorate at some point.
NonFE+CoreFE simulates an extraction method that extracts the same number of formulaic and non-formulaic words.
This should cause lower performance than CoreFE because non-formulaic words are heavily weighted.

\subsection{Baselines for Extraction}

\subsubsection{Phrase Extraction and Sequential Labelling}

We compared our proposed method to other existing methods, which can be classified into two types: phrase extraction and sequential labelling.
For phrase extraction, we adopted LatticeFS \citep{Brooke2017}, a method to extract phrases from a whole corpus.
For sequential labelling \citep{Iwatsuki2018}, each word in a sentence was labelled as either formulaic or non-formulaic.
We adopt two methods: frequency-based and latent Dirichlet Allocation (LDA)-based \citep{Liu2016}.

\subsubsection{LatticeFS}

\citet{Brooke2017} proposed a method (LatticeFS) to extract formulaic expressions by comparing candidate formulaic expressions according to a proposed objective function called \textit{explainedness}.
Their idea is that if one $n$-gram can be explained by another $n$-gram, both can be grouped into one $n$-gram.

They first created an $n$-gram lattice in which the $(n-1)$-gram and $(n+1)$-gram are connected to the $n$-gram.
Then, using the concepts of \textit{covering}, \textit{clearing} and \textit{overlap}, they optimised explainedness to determine which nodes in the lattice should be labelled as formulaic expressions.

\begin{figure}[t]
    \centering
    \includegraphics[scale=0.48]{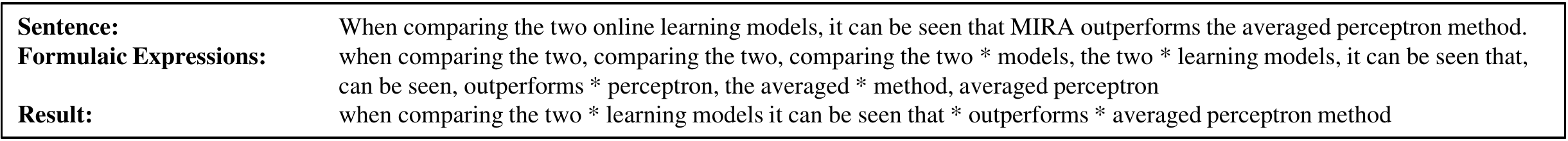}
    \caption{Example of LatticeFS. This method extracts all formulaic expressions from a corpus that are labelled as such by the proposed algorithm. There can be some formulaic expressions that overlap each other.}
    \label{fig:example_Brooke}
\end{figure}

We used the implementation provided by the authors\footnote{\url{https://github.com/julianbrooke/LatticeFS}} and applied it to the FECFeval dataset (for an example, see Figure~\ref{fig:example_Brooke}).
For statistical calculation, a whole corpus is needed and we used AASC.

\subsubsection{Frequency-Based Sequential Labelling}

Formulaic expressions are considered to consist of words that occur more frequently than words that are specific to certain research topics.
According to past work \citep{Iwatsuki2018}, simply removing words with low frequencies improves the performance of classification of communicative functions.

Following this idea, we implemented a frequency-based extraction method consisting of the following steps.
First, we calculated the frequencies of all words occurring in AASC.
Secondly, from a given sentence, we removed all words whose frequencies were lower than the threshold.
In our experiment, we used several thresholds.

\subsubsection{LDA-Based Sequential Labelling}

\citet{Liu2016} applied a topic-modelling to remove unnecessary words from a sentence.
They assumed that words that frequently appear in a certain research topic do not compose formulaic expressions.

They use LDA to assign topic-dependency to each word in a sentence.
They calculated the score that indicates how much a word is a structure word (non-topic word) rather than a topic word as follows:
\begin{eqnarray}
    \mathrm{P}(w) = 1 - \frac{\max p_w(i)}{\sum p_w(i)}, \nonumber
\end{eqnarray}
where $p_w(i)$ is the probability of word $w$ in a topic $i$.

\begin{figure}[t]
    \centering
    \includegraphics[scale=0.48]{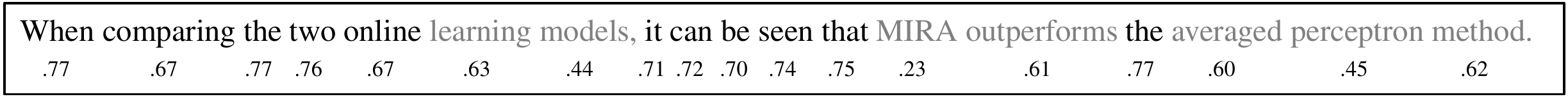}
    \caption{Example that the LDA-based method was applied to. The numbers $\mathrm{P}(w)$ were assigned to each word. Words coloured grey are below the threshold (0.65).}
    \label{fig:LDA}
\end{figure}

Words with $\mathrm{P}(w)$ smaller than the threshold are removed from a sentence.
Following \citet{Liu2016}'s experimental settings, we set the threshold to 0.65 and the number of topics to 10.
The calculation of $\mathrm{P}(w)$ was conducted on AASC.
Figure~\ref{fig:LDA} shows an example.

\section{Results}

\subsection{Validity of the Sentence Retrieval Task as an Extrinsic Evaluation Method}

\begin{figure}[t]
    \centering
    \includegraphics[scale=0.75]{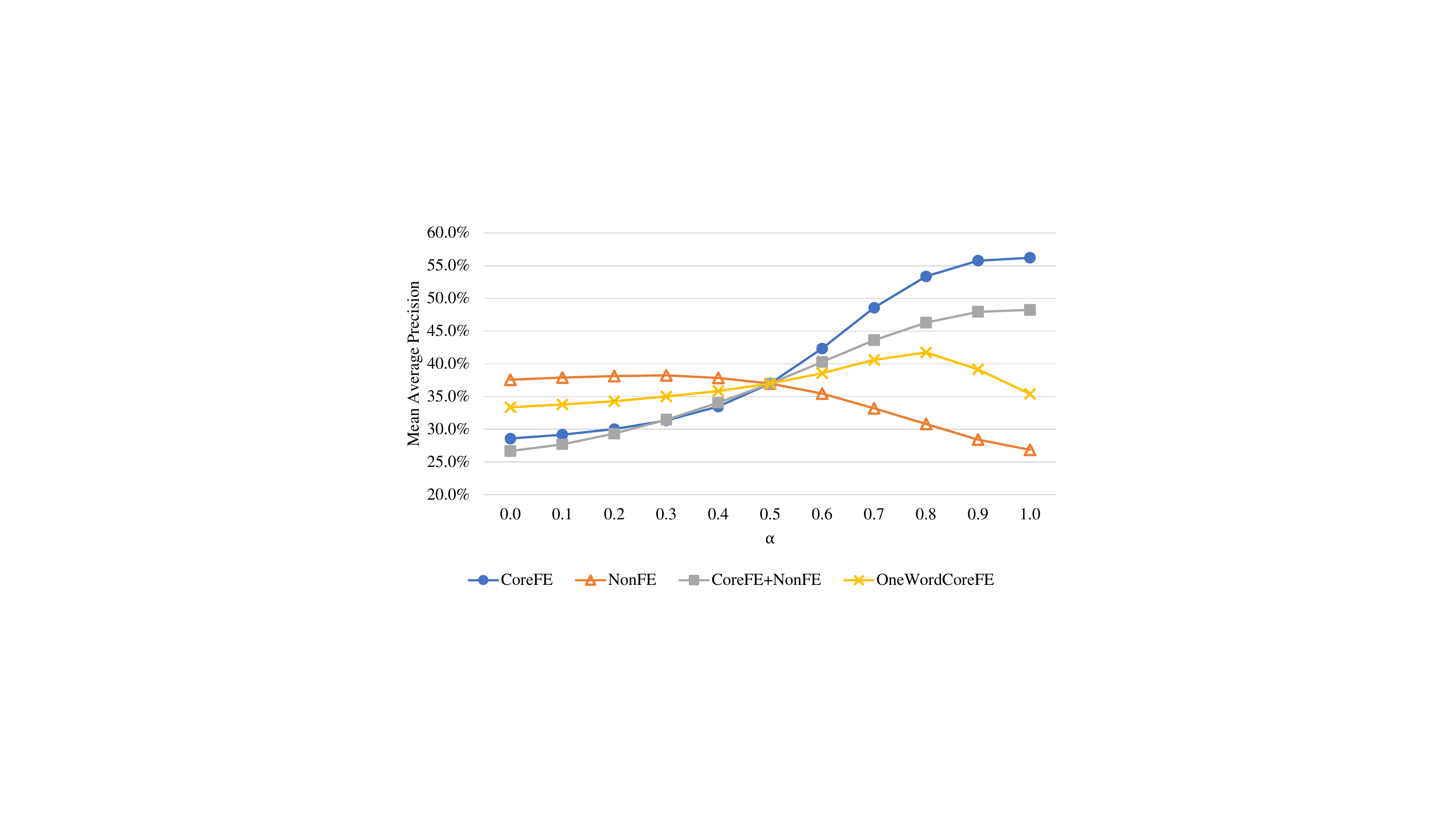}
    \caption{Relationships between MAP and $\alpha$. MAP of CoreFE monotonically increases, while that of NonFE behaves inversely. CoreFE+NonFE shows that lower performance is attributed to extraction of unnecessary words. OneWordCoreFE shows that by missing indispensable words the peak of MAP appears between $\alpha=0.5$ and $1.0$.}
    \label{fig:CoreFE_and_random}
\end{figure}

In Figure~\ref{fig:CoreFE_and_random} the MAP scores of CoreFE, NonFE, CoreFE+NonFE and OneWordCoreFE are shown.
Comparing the performances between CoreFE and NonFE extraction, it can be said that good extraction methods improve the sentence retrieval performance as $\alpha$ increases while bad methods deteriorate the performance as $\alpha$ increases.
Therefore, the MAP score at $\alpha=1.0$ can be used as an indicator of effectiveness of extraction methods.

We conducted further analysis of the transitions of the performances according to $\alpha$.
As for CoreFEs, i.e., good formulaic expressions, MAP increases monotonically as $\alpha$ increases.
Conversely, for NonFE, MAP decreases monotonically.
MAP of CoreFE+NonFE is located between the two.
The performance increases as well as CoreFEs, but due to non-formulaic words, it is not as good as CoreFEs.

However, for OneWordCoreFE, the peak is at, $\alpha=0.8$, and MAP decreases after that.
This phenomenon can be explained as follows.
As $\alpha$ increases from 0.5 to 0.8, heavier weight on the one-word formulaic expressions has a good effect on the performance.
In other words, less weight is put on the remaining formulaic expressions.
This smaller weight on the remaining formulaic expressions deteriorates the performance with higher $\alpha$.

From these observations, we argue that the sentence retrieval task is valid to evaluate extraction methods.
Basically, comparing MAP scores at $\alpha=1.0$ is a good indicator.
The change of MAP score gives additional insight.
If it increases monotonically, most formulaic words are extracted from a sentence.
If there is a peak between $\alpha=0.5$ and $1.0$, the method seems to fail to extract a significant part of a formulaic expression.

\subsection{Formulaic Expression Extraction}

\begin{table}[h]
    \centering
    \caption{Results of all compared methods. The proposed method, named entity removal (NER) and dependency parsing (depparse) achieved the best performance.}
    \begin{tabular}{ccccccc} \hline
         & CoreFE & NonFE & Frequency & LatticeFS & LDA  & NER+depparse \\ \cline{2-7}
        MAP & 56.2\% & 26.9\% & 36.8\% & 35.2\% & 38.6\% & \textbf{42.2\%} \\ \hline
    \end{tabular}
    \label{tab:result}
\end{table}

Table~\ref{tab:result} shows the results of the extraction of formulaic expressions with the proposed and existing methods.
CoreFE and NonFE are also included in the table for comparison.
MAP scores are computed at $\alpha=1.0$.
Among the four extraction methods, the proposed method achieved the best performance.

We also tested various parameter settings for the frequency-based and LDA-based methods to see the differences.
Table~\ref{tab:frequency} shows the MAP scores of the frequency-based method at $\alpha=1.0$ with different thresholds.
Too strict a threshold ($10^{-4}$) seems to remove formulaic words.
There is not much difference between $10^{-5}$ and $10^{-6}$, which implies that almost all words, including formulaic and non-formulaic words, remain as the formulaic part, resulting in the use of whole sentences.

Table~\ref{tab:lda_parameters} shows the MAP scores with different parameters of the LDA-based method.
\citet{Liu2016} reported that based on their experiments, they set the number of topics to 10 and the threshold to 0.65.
This setting is not the best in our experimental settings, but using different parameters did not result in sufficient improvement to outperform our proposed method.

\begin{figure}[t]
    \centering
    \includegraphics[scale=0.48]{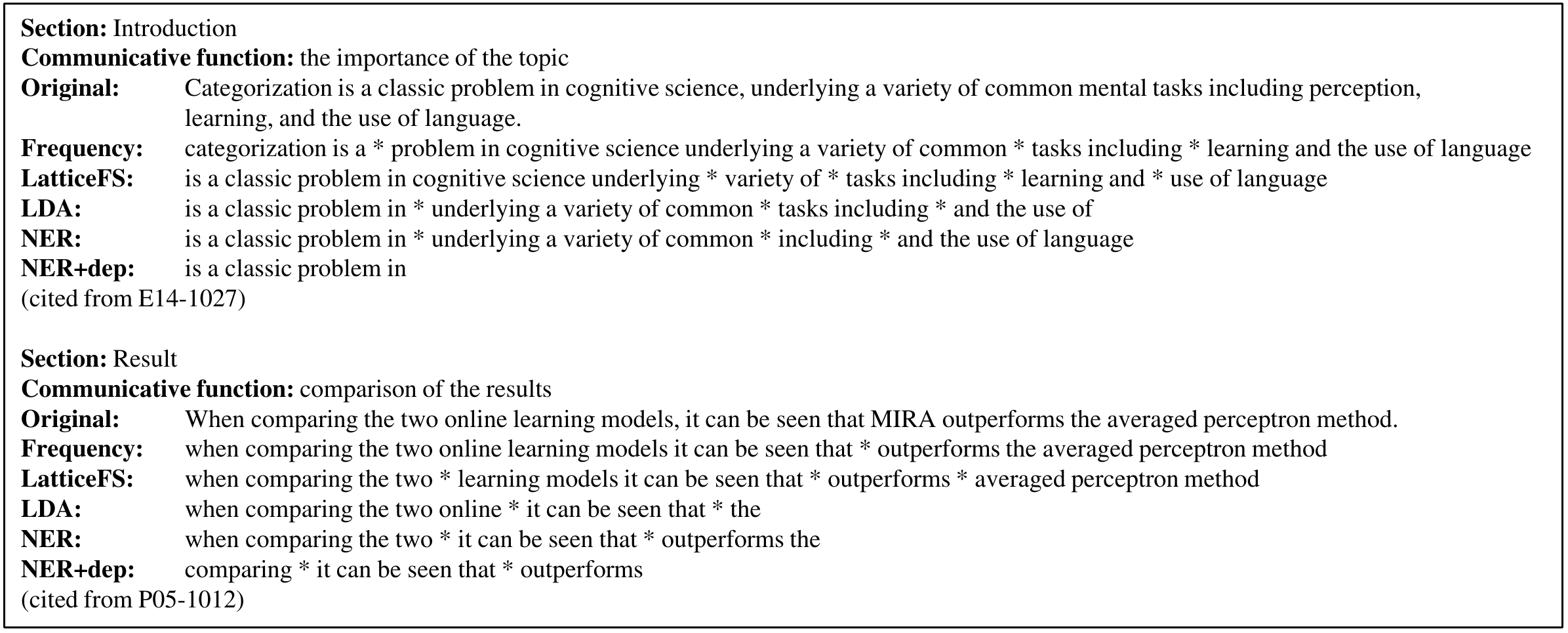}
    \caption{Two examples of results by each method. These sentences were originally retrieved from \citet{Frermann2014} and \citet{McDonald2005}.}
    \label{fig:example_4methods}
\end{figure}

\begin{table}[h]
    \centering
    \caption{Results at $\alpha=1.0$ with different thresholds of frequency.}
    \begin{tabular}{cccccc} \hline
        Threshold & $10^{-4}$ & $5\times10^{-4}$ & $10^{-5}$ & $5\times10^{-5}$ & $10^{-6}$ \\ \hline
        MAP & 35.8\% & 36.8\% & 36.7\% & 36.7\% & 36.6\% \\ \hline
    \end{tabular}
    \label{tab:frequency}
\end{table}

\begin{table}[h]
    \centering
    \caption{MAP scores of LDA-based method with different parameters. Although some combination of parameters achieved relatively low scores, most patterns resulted in no significant difference. We used parameters reported by \citep{Liu2016}, namely 10 topics and 0.65 as the threshold.}
    \begin{tabular}{lcccc} \hline
         & \multicolumn{4}{c}{Number of topics} \\ 
         Threshold & 5 & 10 & 15 & 20 \\ \hline
         0.55 & 39.7\% & 38.6\% & 38.3\% & 38.5\% \\
         0.65 & 36.0\% & 38.6\% & 38.3\% & 38.9\% \\
         0.75 & 31.9\% & 30.3\% & 36.4\% & 39.3\% \\ \hline
    \end{tabular}
\label{tab:lda_parameters}
\end{table}

\section{Discussion}

\begin{figure}[t]
    \centering
    \includegraphics[scale=0.75]{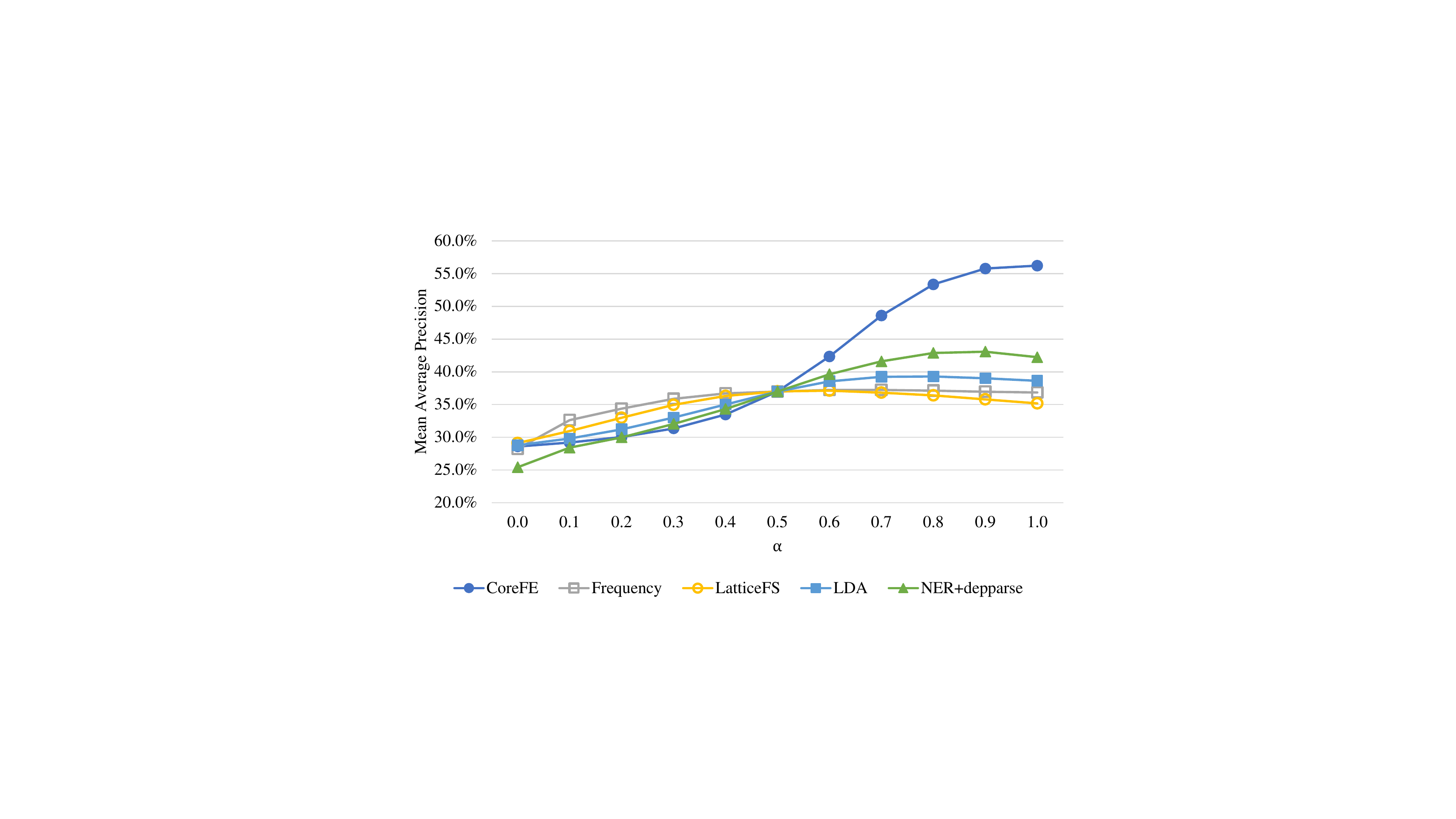}
    \caption{Relationships between MAP and $\alpha$. The peak of the proposed method (NER+depparse) is at $\alpha=0.9$.}
    \label{fig:proposed_method}
\end{figure}

In Figure~\ref{fig:example_4methods}, the formulaic expression candidates extracted by all the methods we tested are depicted.
It was found that the proposed method extracted shorter formulaic expressions than the others did, which implies that it removed non-formulaic words more thoroughly, resulting in better performance.
In Figure~\ref{fig:proposed_method}, the relationships between $\alpha$ and MAP scores are illustrated. 
The peak of the performance of the proposed method, NER+depparse, is at $\alpha=0.9$.
Thus, although it achieved the best performance among other methods, the proposed method missed some formulaic words.

Without dependency-structure-based word selection, the MAP score was 39.8\%, which is higher than that of the LDA-based method (38.6\%) but lower than that of the proposed method (42.2\%).
Therefore, the word selection method worked well to remove non-formulaic words that were not removed by simply applying named entity removal.

We have two types of named entities: general named entities with the CoNLL04 dataset \citep{Roth2004} and scientific entities with the SciERC dataset \citep{Luan2018}.
The MAP score of NER was 39.8\%, but without CoNLL04 dataset, the performance reduced to 39.7\%.
Although the difference was small, it can still be said that both types of named entities worked complementarily.

\section{Conclusion}

There exists a problem that formulaic expressions appear in a sentence with different spans and forms, which has brought difficulty to the extraction and evaluation of formulaic expressions.
To alleviate this problem, we presented the idea that a sentence can be split into a formulaic expression that conveys a communicative function and non-formulaic part that expresses content.
With this approach, formulaic expressions with different spans and forms can be dealt with.
Based on this formulation, we proposed an extraction and evaluation method for formulaic expressions.
Our extraction method consists of named entity removal and dependency structure-based word selection and it achieved the best performance compared to other existing methods.
Our evaluation method adopts the sentence retrieval task as a means of extrinsic evaluation, which measures the strength of the connection between formulaic expression candidates and communicative functions.
We experimentally demonstrated that the proposed evaluation method worked well by evaluating formulaic and non-formulaic expressions.

This work can be utilised to create lists of formulaic expressions automatically, which will accelerate multi-disciplinary academic writing assistance.
We hope that this work will promote research on formulaic expressions in natural language processing and the applied linguistic community. 

\section*{Acknowledgements}

This work was supported by JSPS KAKENHI Grant Numbers 19J12466 and 18H03297 and by Atlanstic 2020 sabbatical grant IKEBANA.

\bibliography{mybibfile}

\end{document}